\documentclass[journal,onecolumn]{IEEEtran}

\usepackage{graphicx}
\graphicspath{ {./images/} }

\usepackage{caption}
\usepackage{amsmath}
\usepackage{float}

\begin{document}

\title{Learning To Play Atari Games Using Dueling Q-Learning and Hebbian Plasticity}

\author{Ashfaq Salehin,~\IEEEmembership{University of Sussex}}

\maketitle

\begin{abstract}

In this work, an advanced deep reinforcement learning architecture is used to train neural network agents playing atari games. Given only the raw game pixels, action space, and reward information, the system can train agents to play any Atari game. At first, this system uses advanced techniques like deep Q-networks and dueling Q-networks to train efficient agents, the same techniques used by DeepMind to train agents that beat human players in Atari games. As an extension, plastic neural networks are used as agents, and their feasibility is analyzed in this scenario. The plasticity implementation was based on backpropagation and the Hebbian update rule. Plastic neural networks have excellent features like lifelong learning after the initial training, which makes them highly suitable in adaptive learning environments. As a new analysis of plasticity in this context, this work might provide valuable insights and direction for future works.

\end{abstract}

\section{Introduction}

\IEEEPARstart{R}einforcement learning is a computational technique where an agent learns by directly interacting with its environment without having a complete model of the environment [1]. Reinforcement learning is a very good example of adaptive systems where an agent learns to make decisions and take actions in an environment in order to maximize some reward, which acts as feedback from the environment to the agent.\\

Well-crafted reinforcement learning agents with optimized training loops are known to learn complex tasks, such as playing computer games. In previous work, a CNN-based agent was trained using discounted policy gradients, where all the rewards in an episode were fed to the agent as training data after discounting by a factor [2]. Although this approach served as a good starting point, it is not suitable for learning to control complex environments, such as Atari games.\\

A better implementation is possible using the Q-Learning algorithm, which is based on the Bellman equation [3]. The Bellman equation is based on the Markov decision process [4] and states that the optimal value of a state is equal to the immediate reward plus the discounted expected optimal value of the next state under the optimal policy. While the Bellman equation requires all the reward values and transition probabilities to be known in advance, the Q-Learning algorithm uses Q-Values, which are initialized as random values and optimized gradually. Deep Q-Learning (DQN) is another variant of Q-Learning that uses a neural network agent responsible for predicting the optimal Q-values for any given state and improving its estimation over time.\\

Furthermore, basic DQN is not optimal for complex environments, as the system can take a very long time to learn something useful. Researchers have proposed various strategies to mitigate this issue, such as Double DQN, where two separate neural networks are used for providing the Q-value targets and finding the best action [4], and Dueling DQN, where the agent learns not only the optimized action but also the "advantage" of taking the action over others [5]. These optimization techniques can be combined for even better results, and in this work, a Dueling Double DQN is implemented and compared with a Double DQN implementation.\\

Until this point, the work is a simplified reimplementation of DeepMind’s work [6] and [7]. The reimplementation was necessary because their code is not open-sourced. Some third-party implementations can be found that are neither documented nor easily transferable. An original attempt was made to use a plastic neural network-based agent in combination with Dueling Double DQN.\\

Plastic neural networks have two different sets of weights: the fixed weights and the plastic weights. After the initial training, the fixed weights are frozen, and the plastic weights are left open for training. The final output from the network is a weighed sum of the fixed and plastic weights. Such an architecture lets plastic neural networks learn from real experience and allows lifelong learning. In this work, the paper about differential plasticity from Uber research [9] is implemented and integrated with Double and Dueling DQN implementations.\\

Finally, the rewards from the plasticity-based agent are thoroughly compared with the raw Dueling Double DQN rewards over time. This work has the potential to give direction to the growing body of research on lifelong learning and the application of neuroplastic models to reinforcement learning problems.

\section{Background}

Reinforcement learning, initially referred to as "optimal control," has been around since the early 1950s. It was developed to minimize the measure of a dynamic system's behavior over time [10]. Arthur Samuel first utilized this concept in 1959 in the area of artificial intelligence. He employed a form of reinforcement learning to enhance the strategy of a checkers-playing program progressively over time [11].\\

In his famous 1957 paper [3], Richard Bellman introduced the mathematical framework of Markov decision processes, which model sequential decision-making problems where an agent interacts with an environment. This mathematical equation, namely the Bellman optimality equation, can be used to find the optimal value of any given state. There is a problem with the Bellman optimality equation; it requires all the state transition probabilities and reward values to be known in advance, which is not always possible. To mitigate this, Watkins and Dayan [4] introduced Q-learning. Without requiring full knowledge of the system, the Q-learning learns an action-value function called the Q-function, which estimates the expected future reward for taking a certain action in a given state. The Q-learning update rule allows the agent to iteratively update the Q-values based on the rewards received and the estimated future rewards, eventually converging to the optimal Q-function. 
The field of reinforcement learning took a significant leap forward when a team from DeepMind developed a system that could learn to play any Atari game from scratch, solely by processing the raw screen pixels, without being explicitly programmed with the rules of the game [7]. The researchers continued to refine their algorithm, enabling it to surpass human-level performance in those Atari games [8]. Another groundbreaking achievement occurred when the program created by Silver et al. defeated the reigning world champion in the game of Go [12].\\

DeepMind’s research on Atari game-playing agents is pioneering because they first successfully used Deep Q-Leaning. Deep Q-learning is a variation of the original Q-learning algorithm where a neural network is used to predict the optimal Q values. The Q-value prediction network is trained using gradient descent, where the model is penalized with the difference between the predicted and the target Q values. DeepMind’s first paper was solely based on a plain DQN implementation without much optimization. But their later papers were based on advanced DQN implementations like Double DQN [5], Duelling DQN [6], and other optimizations like prioritized experience replay [13]. In their latest paper, DeepMind has combined six of these optimization techniques to obtain superior performance [14].\\

This work also explored integrating plasticity with the Deep Q-Network (DQN) approach in the context of Atari game-playing agents. The concept of plasticity was originally proposed by Donald Hebb in his seminal 1949 paper [15], which laid the foundations for many models of synaptic plasticity in neural networks. In recent years, a research team from Uber has revisited the idea of Hebbian plasticity in the context of neural networks [9]. In their implementation, the underlying model has two distinct sets of weights: fixed weights and plastic weights (also known as Hebbian traces). The fixed weights are trained using a standard neural network training loop, while the plastic weights are also trained using backpropagation but are additionally updated according to the Hebbian update rule.\\

The idea of using reinforcement learning and DQN to master an environment came from reading the book written by Aurelien Geron [16]. In this book, the author used raw, double, and duelling DQNs to maximize the rewards in the CartPole environment. Whether the explanations were very helpful in understanding the concept, the idea of using the same techniques in the Atari environments came from further literature review. The idea of using synaptic plasticity in this context is an original one that came from an exploration of related literature.\\

\section{Deep Reinforcement Learning}

\subsection{Concept}

In reinforcement learning terminology, an agent interacts with the environment by performing some action. The environment represents the system or problem domain that the agent is trying to navigate and control. The environment maintains it’s internal states, which get updated with actions performed by the agent. The agent's goal is to learn an optimal policy - a mapping from the current state of the environment to the best action to take. The policy takes as input the reward value generated from the previous action and the environment’s current state. After receiving these inputs, the policy updates it’s internal mapping and returns the likelihood of the next action among all possible actions. From this likelihood, the next action is selected and applied to the environment. The environment state gets updated, the new reward value gets returned, and then the process is repeated. If it runs for a sufficiently large number of iterations, the system will learn to maximize the rewards obtained in the environment.\\

A block diagram will help to understand the workflow better.

\begin{figure}[h]
    \centering
    \includegraphics[scale=0.5]{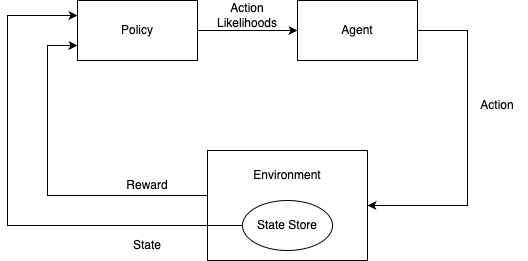}
    \caption{Reinforcement Learning System Diagram}
\end{figure}

\subsection{Bellman Optimality Equation}

In the early 20th century, Andrey Markov developed a memoryless stochastic process, namely the Markov chain, which Bellman studied in his 1957 paper [3]. Markov processes have a fixed number of states, and the transition probabilities between these states are fixed. The probability of any state transition is associated with the destination state $s$ and the previous state $s’$, and none else. Transition values can be associated with rewards, which can be either negative or positive.

\begin{figure}[h]
    \centering
    \includegraphics[scale=0.5]{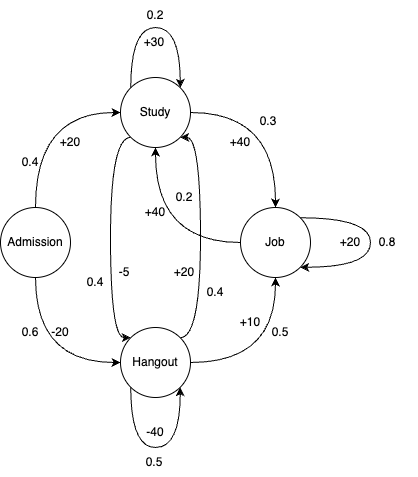}
    \caption{An Example of Markov Decision Process}
\end{figure}

The Markov process can be explained with the fun example illustrated in the above figure. After admission to university, many students spend their time hanging out and doing other recreational activities, harming their academic progress, which can be thought of as a negative reward. On the other hand, some students study attentively, resulting in good academic grades, which can be thought of as a positive reward. Continuous study brings even better results, hence a higher reward. But some students get bored and go for recreational activities, which negatively affects their studies at a certain level. Continuously hanging out may fail the student; hence, the reward value is highly negative, but returning to study can still bring good grades, hence positive rewards. Attentive students quickly get jobs after finishing their studies, where the reward value is very high. On the contrary, students who spent most of their time hanging out might also get jobs, but with a lower probability and a lower salary, hence a lower reward value. Most people who get jobs will hop jobs and never return to study. They will generally grow slower, hence the low reward value. On the other hand, some students will return for higher studies with experience, do impactful research, and hence get a very high reward.\\

Richard Bellman came up with an equation in 1957 named the Bellman Optimality Equation, which serves as the foundation of reinforcement learning. According to the equation, the accumulated optimal rewards in the current state $s$ is [3],

\begin{equation}
\label{eq:bellman_optimality}
V^*(s) = \max_a \sum_{s'} T(s, a, s')\left[R(s, a, s') + \gamma \cdot V^*(s')\right]
\end{equation}

\[
\begin{aligned}
&\text{Where,}\\
&s \text{ - Current state}\\
&s' \text{ - Next state}\\
&a \text{ - The action to take in the current state}\\
&a' \text{ - The action which will be taken in the next state}\\
&T(s, a, s') \text{ - The probability of going to $s'$ from $s$, given that the chosen action was $a$}\\
&R(s, a, s') \text{ - The reward obtained through the transition to $s'$ from $s$, taking the action $a$}\\
&\gamma \text{ - The discount factor, this is a factor to credit a series of actions for the obtained reward}\\
&V^*(s') \text{ - The optimal value of the destination state $s'$, given the fact that the agent will play optimally in the future}
\end{aligned}
\]

In words, the optimal accumulated reward at any state $s$, denoted by $V^*(s)$, is the maximum value among all possible actions $a$ in this state, where the value for each action is computed as the sum of two terms:

\begin{enumerate}
\item The immediate reward $R(s, a, s')$ received for taking action $a$ in state $s$ leading to state $s'$.
\item The discounted optimal accumulated reward for the next state $s'$, weighted by the transition probability $T(s, a, s')$ of reaching $s'$ from $s$ by taking action $a$.
\end{enumerate}

\vspace{0.3cm}

The Bellman optimality equation can be converted to a recursive form known as the value iteration algorithm [16].\\

\begin{equation}
\label{eq:value_iteration_algorithm}
V_{k + 1}(s) = \max_a \sum_{s'} T(s, a, s')\left[R(s, a, s') + \gamma \cdot V_k(s')\right]
\end{equation}

\begin{flalign*}
&\text{All the notations are same as the Bellman optimality equations, except,} & \\
&V_{k + 1}(s) \text{ - The optimal accumulated reward value in state $s$ at the current time step.} & \\
&V_k(s') \text{ - The optimal accumulated reward value in state $s'$ at the next time step.} &
\end{flalign*}

While knowing the optimal state value function $V^*(s)$ is useful, it is not sufficient to determine the optimal policy for an agent, as the agent needs to know the optimal action to take in each state. Computing the optimal value function for all possible actions in every state can be time-consuming. Instead, it is more beneficial to compute the optimal state-action value function, also known as the Q-function or Q-values [3].\\

\begin{equation}
\label{eq:q-value-algorithm}
Q_{k + 1}(s, a) = \sum_{s'} T(s, a, s')\left[R(s, a, s') + \gamma \cdot \max_{a'}Q_k(s', a')\right]
\end{equation}

\begin{flalign*}
&\text{The notation is similar, except,} & \\
&Q_{k + 1}(s, a) \text{ - Optimal state-action or Q-value for state action pair $(s, a)$ at current time step.} &\\
&Q_{k}(s', a') \text{ - Optimal state-action or Q-value for state action pair $(s', a')$ at the next time step.}
\end{flalign*}

The value iteration algorithm (Eq \ref{eq:value_iteration_algorithm}) and the Q-value algorithm (Eq \ref{eq:q-value-algorithm}) allow to compute the optimal reward value at any state using a dynamic algorithm implementation. But an important thing to notice is that these algorithms require the system properties (transition probabilities, reward values, etc.) to be known in advance.\\

\subsection{Q-Learning}

Q-Learning is an adaptation of the Q-value optimization algorithm (Eq \ref{eq:q-value-algorithm}) for systems where the reward values and the state transition probabilities are initially unknown. The system starts with a random set of Q values. Then the system “trains” by watching an agent play and gradually improving the Q-value estimates. The degree by which the Q-values are updated is called the "learning rate". Once a predefined number of episodes has been passed or the Q-value estimates are close enough, then the system can select the best action by selecting the one with the highest Q-value estimate.\\

The Q-learning algorithm can be represented by the following equation [16].\\

\begin{equation}
\label{eq:q-learning-equation}
Q(s, a) = \alpha r + \gamma \cdot \max_{a'} Q(s', a')
\end{equation}

\begin{flalign*}
&\text{In the Q-learning equation,} & \\
&Q(s, a) \text{ - The Q-value for $(s, a)$ state action pair at current time step.} & \\
&Q(s', a') \text{ - The Q-value for $(s', a')$ state action pair at the next time step.} & \\
&r \text{ - The reward value obtained transitioning from state $s$ to state $s'$ through action $a$.} & \\
&\alpha \text{ - The learning rate, how much of Q-values will be updated on each iteration.} & \\
&\gamma \text{ - The discount factor.}
\end{flalign*}

\subsection{Deep Q-Learning}

Q-learning algorithm can learn the Q-values without requiring state transition probabilities, but it needs to compute the Q-values for all possible states. For an environment with a large action space, computing the Q-values of all possible states is impractical and somehow impossible. An alternative way is to learn a function $Q_\theta(s, a)$ that can find the approximate Q-values of any state-action pair with a manageable number of parameters.\\

Neural networks can be used to efficiently approximate the Q values without doing any feature engineering, which also works much better, as shown by DeepMind in 2013. The result is an architecture called Deep Q-Networks, or DQN. Deep Q networks also use a modified version of the Bellman equation. The target Q-value in the DQN algorithm is,\\

\begin{equation}
\label{eq:deep-q-learning}
y(s, a) = r + \gamma \cdot \max_{a'} Q_\theta(s', a')
\end{equation}

In deep Q-learning, an exploration-exploitation strategy known as the epsilon-greedy policy is employed during training. Initially, the agent predominantly explores the environment by taking random actions, allowing it to gather diverse experiences and discover potentially rewarding actions. As training progresses, the agent gradually shifts towards exploiting the learned Q-values by selecting actions with the highest estimated Q-values more frequently, refining its policy towards optimal behavior. However, a small probability of taking random actions is maintained throughout the training process to ensure continued exploration, adaptation to changes, and the potential discovery of better policies. This balance between exploration and exploitation is controlled by a decreasing epsilon value. Taking the exploration policy into account, the action a is,

\begin{equation}
\label{eq:deep-q-learning-with-exploration-policy}
\begin{aligned}
a = \begin{cases}
    a_{\text{rand}} & \text{if } \text{rand} < \epsilon \\
    \arg\max_{a'} Q_\theta(s', a') & \text{if } \text{rand} \geq \epsilon
\end{cases}
\end{aligned}
\end{equation}\\

The loss between the target and predicted Q-values can be computed using,

\begin{equation}
\label{eq:dqn_loss}
L(\theta) = \frac{1}{N} \sum_{i=1}^{N} \left( y(s_i, a_i) - Q_\theta(s_i, a_i) \right)^2
\end{equation}\\

The model parameters are updated with the gradients of the loss against the trainable variables.

\begin{equation}
\label{eq:dqn_update}
\theta \leftarrow \theta - \alpha \cdot \nabla_\theta L(\theta)
\end{equation}

\begin{flalign*}
&\text{In the DQN equations,} & \\
&y(s, a) \text{ - Target Q-value for the state-action pair $(s, a)$}. & \\
&Q_\theta(s, a) \text{ - Estimated Q-values for the current state-action pair $(s, a)$.} & \\
&Q_\theta(s', a') \text{ - Estimated Q-values for the next state-action pair $(s', a')$.} & \\
&\epsilon \text{ - Controlling variable for epsilon-greedy policy.} & \\
&r \text{ - The reward obtained by the state-action pair $(s, a)$.} & \\
&\gamma \text{ - The discount factor.} & \\
&rand \text{ - A random value.} & \\
&\theta \text{ - The model parameters.} & \\
&L(\theta) \text{ - The loss function which depends on the parameters of the network.} & \\
&N \text{ - The batch size.}
\end{flalign*}

Another important component of deep Q-learning is the experience-replay buffer. This buffer is typically large enough to store a considerable number of experiences encountered during training. Various eviction policies can be employed, such as FIFO (First-In-First-Out) queue combined with random sampling, or a priority queue with priority-based sampling and eviction. By randomly sampling batches of experiences from this buffer during training, the model learns from a diverse set of past experiences, mitigating issues of correlated data and facilitating more stable learning. Alternatively, the experience-replay buffer can sample important experiences with the highest TD error (loss), to retry the experiences the model is most confused with [13].

\subsection{Deep Q-Learning Variants}\label{subsec:dueling-dqn}

There are several variants of deep Q-learning. If the same network is used to find the target Q-values and predict the current Q-values, a feedback loop occurs, which can make the training unstable. To fix this, two DQN’s are used, one for each purpose. The target setting model is just a copy of the original model, and it’s weights are synced with the main model after every fixed number of episodes. Because the target setting model is updated less frequently, the targets are more stable, and the overall training stability increases. In another variant, the main model is used to find the best actions for the next step, and the target model is just used to estimate the Q-values of these best actions. This architecture is called Double DQN [5].\\

In another variant, an advanced model architecture is used. It was developed based on the idea that the Q-value of a state-action pair $(s, a)$ can be represented as the sum of the value of the state $V(s)$ and the advantage of taking the action $a$ in the state $s$, $A(s, a)$.

\begin{equation}
\label{eq:dueling-dqn}
Q(s, a) = V(s) + A(s, a)
\end{equation}

The value of a state is equal to the Q-value of it's best action $a*$.

\begin{equation}
\label{eq:dueling-dqn-state-value}
V(s) = Q(s, a*)
\end{equation}

This implies the fact that $A(s, a*) = 0$. In this model, the state value and the advantages are both computed, and the best advantage is subtracted from the computed advantages because of the aforementioned fact. This variation is called Dueling DQN [6]. 

\section{Plasticity}

\subsection{Concept}

Plasticity refers to the ability of connections, or synapses, between neurons in a neural network to change their strengths or weights over time. More specifically, plasticity is the process by which the connection strengths are modified based on the activity patterns of the neurons in the network. Different authors have mentioned different ways to implement plasticity, as Miconi et al. [9] from Uber Research proposed that the plasticity of each connection be made trainable, allowing the network to learn not just the weights of connections but also how plastic (adaptable) each connection should be. This is achieved by having two sets of parameters for each connection: a fixed weight and a plasticity coefficient that modulates a Hebbian trace. The plasticity coefficients and fixed weights are optimized through backpropagation to maximize performance on the learning task, essentially automating the discovery of an efficient plasticity mechanism for this task. The key idea is that instead of hand-designing plasticity rules, the plasticity mechanism itself can be learned in a similar way as the ordinary parameters. Plastic neural networks with trainable plasticity mechanisms support lifelong learning by enabling the networks to continually acquire and store new information from ongoing experiences, similarly to the learning capabilities of biological neural networks in the brain.

\subsection{Hebbian Plasticity and Backpropagation}

To implement plasticity, a rule must be defined to separate the fixed weights from the plastic weights. Different authors proposed different implementations [17, 18]. Miconi et al. [9] proposed to implement plastic connections as a Hebbian trace matrix $Hebb$, which is initialized with zeros and updated with the Hebbian update rules.The output of each layer is a weighted combination of the fixed and the plastic weights.

\begin{equation}
\label{eq:plastic-layer-output}
x_j(t) = \sigma\{\sum_{i \in inputs} \left[ w_{i,j} x_i(t - 1) + \alpha Hebb_{i, j}(t) x_i(t - 1) \right] \}
\end{equation}\\

The plastic weights are updated with the Hebbian update rules [15],

\begin{equation}
\label{eq:hebbian-weight-updates}
Hebb_{i, j}(t + 1) = \eta x_i(t - 1) x_j(t) + (1 - \eta) Hebb_{i, j}(t)
\end{equation}

\begin{flalign*}
&\text{In the plasticity equations,} & \\
&x_j(t) \text{ - Output of current $j^{th}$ layer, in the current timestep $t$.} & \\
&x_i(t - 1) \text{ - Output of the previous $j^{th}$ layer, in the previous timestep $t - 1$.} & \\
&w_{i, j} \text{ - The $(i, j)^{th}$ component of the weight matrix.} & \\
&Hebb_{i, j}(t + 1) \text{ - The $(i, j)^{th}$ component of the Hebbian matrix in the next timestep $t + 1$.} & \\
&Hebb_{i, j}(t) \text{ - The $(i, j)^{th}$ component of the Hebbian matrix in the current timestep $t$.} & \\
&\sigma \text{ - The activation function of the neural network layer.} & \\
&\alpha \text{ - A parameter that controls the fraction of plastic weights being added to the output of layer $j$.} & \\
&\text{Usually, plastic weights will contribute to the layer output by a small amount.} & \\
&\eta \text{ - A parameter that controls the amount of update in the plastic weights in each iteration.} & \\
&\text{It is a parameter similar to the learning rate, but for plastic weights.}
\end{flalign*}

The plastic weights ($Hebb$ matrices) also gets updated using backpropagation (Eq \ref{eq:dqn_update}).

\section{Implementation}

\subsection{The Environment}

The experiments in this work uses the Atari environments in OpenAI Gym, created by the Arcade Learning Environment (ALE) [19]. Creating an instance of any ALE environment in OpenAI Gym is easy. The “make” method needs to be called on the “gym” object with the correct environment name, which creates the environment object. A seed can be passed while creating the environment to make the experiments reproducible. Calling the reset function on the environment object resets the game. Finally the step function performs an action on the environment and returns the new state, obtained reward value and if the game is ended along with other information. The experiments were run on four atari games, Space Invaders, Breakout, Sea Quest and Beam Rider. But just by passing any environment name, this system can be trained to learn that atari game from scratch.

\begin{figure}[h]
    \centering
    \includegraphics[scale=1.0]{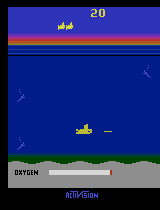}
    \caption{A sample frame from the SeaQuest environment.}
\end{figure}

\subsection{Preprocessing}

The game frames are preprocessed before they are ready to be fed to the neural networks. The frames are converted from color to grayscale images and then the playable area is cropped to obtain $160x160$ images. The cropped frames are then downsized to obtain $84x84$ images. Four of these consecutive states are stacked together to create a history as this helps to predict the Q-values given the dynamic states. To elaborate, the last three states with the current state form the state history and the last two states, the current state and the next state form the next state history. The current and next state history, the reward information and if the episode is stopped or truncated forms an experience and then it is added to the experience replay buffer. Once the experience buffer has sufficient number of experiences, random batches are drawn from the buffer and neural networks are trained with these batches.

\subsection{Neural Network Agents}

In this work, three kind of neural network agents were implemented.\\

\subsubsection{Double Q Network}

The plain DQN agent was implemented as a neural network used for image classification [16]. In this architecture a few convolutional layers are stacked, each followed with a Relu activation layer and a pooling layer. The convolution filters are used to extract features from the network hierarchically. The top layers learn the smaller features in the image while the bottom layers learn the larger features. As the images progress through the network they become smaller and smaller and at the same time gets deeper and deeper. On top of the CNN layers a feedforward network is added composed of a few dense layers with relu activations, dropout layers to prevent overfitting and a final dense layer to output the predictions. In this architecture the outputs produced by the final layer are used as the Q-values.

\begin{figure}[h]
    \centering
    \includegraphics[scale=0.55]{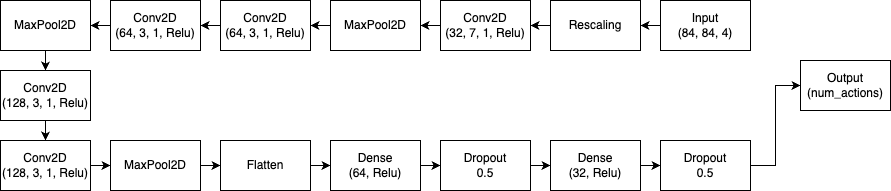}
    \caption{The neural network architecture for the normal double Q networks.}
\end{figure}

\subsubsection{Dueling Q Network}

The Dueling Q-network is implemented based on the description in subsection \ref{subsec:dueling-dqn} and the equations \ref{eq:dueling-dqn} and \ref{eq:dueling-dqn-state-value}. After rescaling, the image samples are passed through a stack of convolutional layers making the images smaller and deeper. The convolutional layers can be thought of as an image encoder. On top of the last convolutional layer a flattening layer is used to flatten the images in a single dimension, followed by a dense layer. The dense layer is fed to two dense layers creating two branches, the state values and the raw advantages. The advantage is normalized by subtracting the max advantage from each raw advantage values. Finally, the state values and the advantages are added to output the Q-values.

\begin{figure}[h]
    \centering
    \includegraphics[scale=0.52]{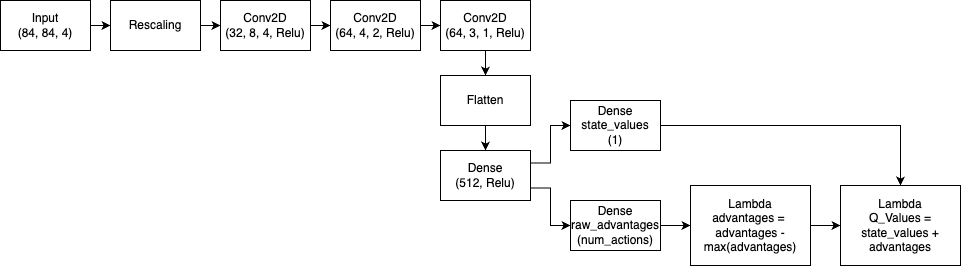}
    \caption{The neural network architecture for the dueling DQN.}
\end{figure}

\subsubsection{Dueling Q Network with Injected Plasticity}

Finally, Hebbian plasticity is used to convert the dueling DQN network a plastic neural network. The convolutional encoding architecture is kept the same, as well as the connections between the later dense layers. The change is in the fact that plastic weights are added between every connecting pair of flatten and dense layers.

\begin{figure}[h]
    \centering
    \includegraphics[scale=0.47]{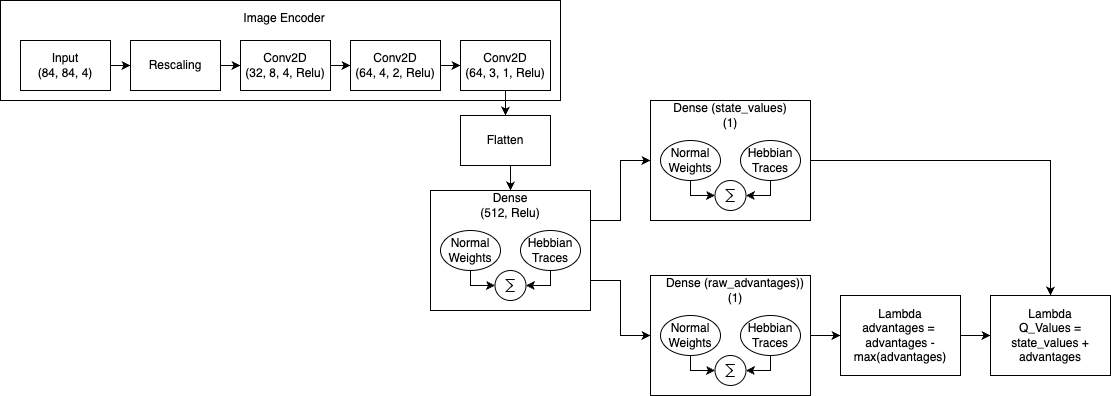}
    \caption{The neural network architecture for the dueling DQN with plasticity injection.}
\end{figure}

\subsection{Training Loop}

\subsubsection{Loss Function and Optimizer}

The training algorithms in this paper uses mean squared error (MSE) as the loss function. This loss function contributes less for small errors but much for large errors to the loss. A Nadam optimizer is used to optimize the model parameters. A learning rate scheduler is used to decay the learning rate from $10^{-2}$ to $10^{-4}$ over 60\% of total episodes. This helps the training algorithm to take bigger steps in the beginning and gradually take smaller steps when it is close to optimization.\\

\subsubsection{Epsilon-Greedy Policy}

The training loops in this paper uses epsilon-greedy policy. The value of $\epsilon$ is annealed from $1.0$ to $0.1$ over the the period of all episodes. A random value is compared with $\epsilon$, if it is less than $eta$ then a random action is taken and if it is greater then the mostly likely action with the highest Q-value is used.\\

\subsubsection{Experience Replay}

The experiments uses a FIFO queue as the replay buffer. A buffer with a maximum size of 50,000 experiences is used and the oldest experiences are removed once the buffer is out of it’s capacity. The buffer object contains a method to sample a random batch from the current elements in the buffer. Ideally a priority buffer should be implemented based on the TD error [12] to focus training on the experiences the model is most confused with. But due to the time limitation, a FIFO replay buffer is used.\\

\subsubsection{Models Without Plasticity}\label{subsec:model-training-loop}

Given all the components in place, the training loop is simple. In this work, all different atari environments are trained for 10,000 episodes, where in each episode a maximum of 3,000 steps are run. In each step an action is sampled using epsilon-greedy policy and the experience is added to the experience replay buffer. After the initial warmup period of fifty episodes, at the end of each episode a random batch is sampled from the replay buffer and the target next Q-values are obtained using equation \ref{eq:deep-q-learning}. The network is then used to predict the Q-values given the current states and the loss is computed using equation \ref{eq:dqn_loss}. Finally, the model parameters are updated using gradient descent according to equation \ref{eq:dqn_update}. The values of total rewards, losses and the maximum Q-value is recorded throughout the training period.\\

\subsubsection{Models With Injected Plasticity}

In the models with plasticity injection, the first 70\% of the episodes are trained according to subsection \ref{subsec:model-training-loop}. Then the fixed weights are frozen and the Hebbian traces are initialized with zeroes. In the final 30\% of the episodes the plastic weights are updated using the Hebbian update rules (equation \ref{eq:hebbian-weight-updates}) and trained using backpropagation. The plasticity learning rate or $\epsilon$ parameter is set to $10^{-3}$ and the plasticity contribution parameter $\alpha$ is set to $0.2$ or 20\%. The outputs from the plastic layers are computed using equation \ref{eq:plastic-layer-output}. The goal of plasticity training is that the agent will learn through real-life experiences. Hence, no experience replay buffer is used in this stage. Instead, the current frames are stacked together to form a state history, and only the experiences from the current episode are sampled to form a batch.

\section{Result and Analysis}

\subsection{Results}

\begin{figure}[h]
    \centering
    \includegraphics[scale=0.7]{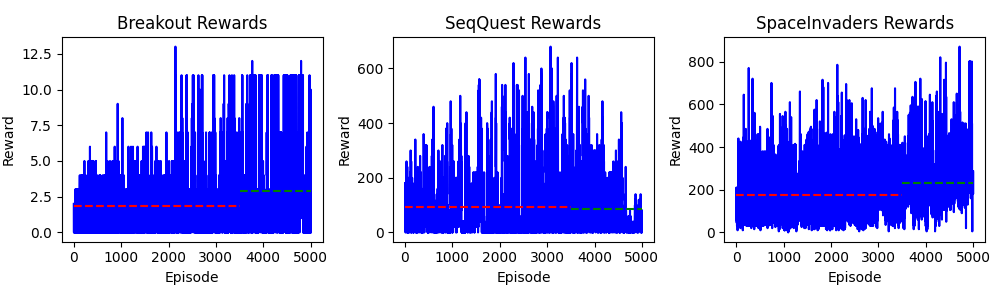}
    \caption{Rewards obtained for Breakout, SeaQuest and SpaceInvaders using raw DQN. The training loop was run for 5,000 episodes. The learning rate was fixed at $10^{-4}$. The dashed lines are average rewards, which are computed for first 70\% and last 30\% episodes separately.}
    \label{fig:dqn_rewards}
\end{figure}

Figure \ref{fig:dqn_rewards} shows the plots of reward values obtained over episodes while trained using raw DQNs. The training loops were run for 5,000 episodes with fixed learning rates. Through the graphs, the reward values are increasing and decreasing after random time intervals. Except for SeaQuest, the higher reward values are noticed toward the ends, hinting that the systems have learned something. For SeaQuest, the reward values fell drastically toward the end, which might be due to the fact that the model was overfitted. For all of the models, the starting values are lower due to random actions being taken, and the values generally increase over time. This is also evident from the fact that the average values of the ending episodes are higher than the averages of the beginning episodes.

\begin{figure}[h]
    \centering
    \includegraphics[scale=0.7]{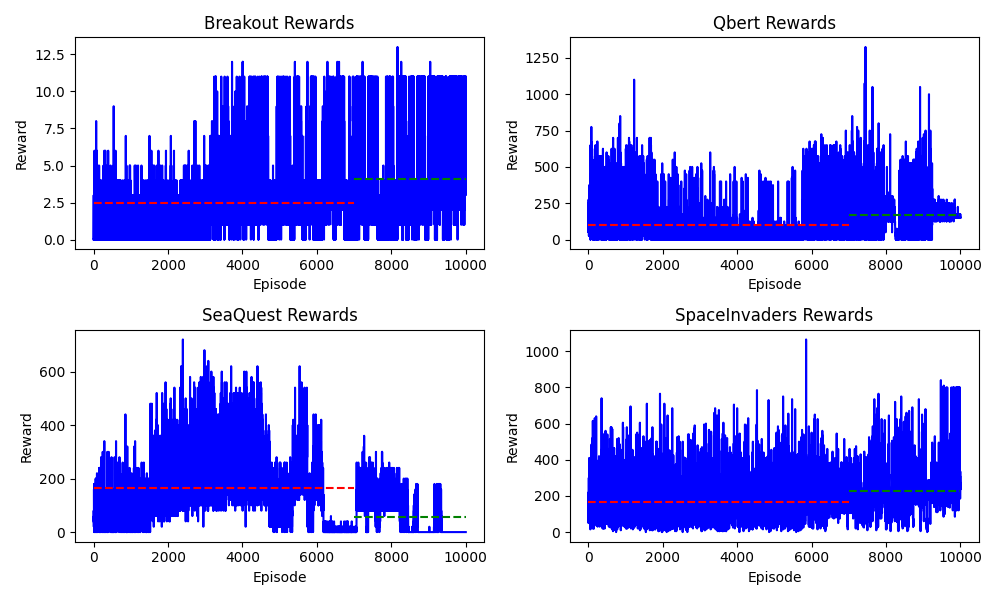}
    \caption{Rewards obtained for Breakout, Qbert, SeaQuest and SpaceInvaders using Dueling DQN. The training loop was run for 10,000 episodes. A learning rate scheduler was used. The dashed lines are average rewards, which are computed for first 70\% and last 30\% episodes separately.}
    \label{fig:ddqn_rewards}
\end{figure}

Next, the plots shown in Figure \ref{fig:ddqn_rewards} show the reward values while the models were trained using dueling DQNs. In these cases, the training was done for 10,000 episodes with a decreasing learning rate from $10^-2$ to $10^-4$. The maximum reward values are higher than raw DQNs in all of these plots. The higher values are consistent toward the ends, except for Qbert and SeaQuest, which again hint at overfitting. Instead of improving due to longer training, the SeaQuest model is more badly overfitted than raw DQN. Other than this, the average rewards toward the ends are higher than the beginning episodes.

\begin{figure}[h]
    \centering
    \includegraphics[scale=0.6]{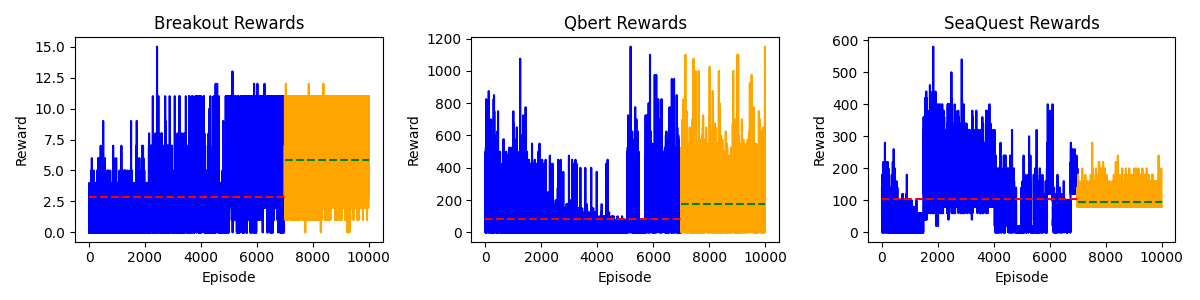}
    \caption{Rewards obtained for Breakout, Qbert, SeaQuest and SpaceInvaders using Dueling DQN with plasticity injection. The training loop was run for 10,000 episodes. A learning rate scheduler was used. The blue part is the reward value obtained from fixed training, while the yellow part is the reward value obtained from plastic training. During the plastic training, an $\epsilon$ value of 0.1 was used. The dashed lines are average rewards, which are computed for the fixed training rewards and the plastic training rewards separately.}
    \label{fig:ddqn_with_plasticity_rewards}
\end{figure}

Interesting results are obtained when the DDQN models are trained with plasticity injections. This was illustrated in Figure \ref{fig:ddqn_with_plasticity_rewards}. The systems were trained for 10,000 total episodes, while in the first 7,000 episodes, the fixed weights were trained, and throughout the rest, only the plastic weights were trained. The reward values during the plasticity training episodes are very consistent and high. For Breakout, the plastic rewards look like a higher band than the fixed training rewards, very rarely falling to zeros. For Qbert, the falls are not very prominent compared to the beginning, and the values are higher. For SeaQuest, the highest value fell in the plasticity phase, but the values are consistent, compared to both raw and dueling DQNs, where the end values are mostly zeros. Even after overfitting, the model for SeaQuest maintained a reasonable performance in the plastic training phase. Moreover, the average reward values are higher in the plastic training phase than in the fixed training phase, and the differences are much more prominent than in the non-plastic networks.

\begin{figure}[h]
    \centering
    \includegraphics[scale=0.8]{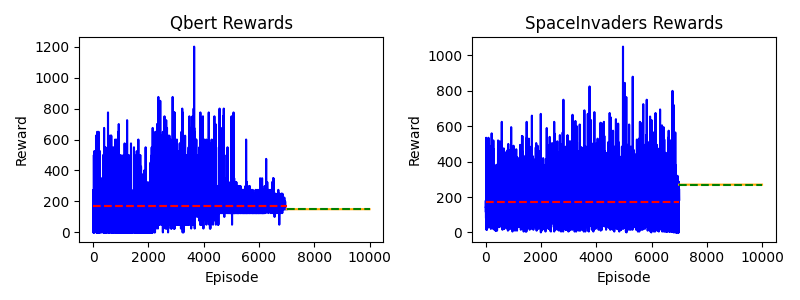}
    \caption{Rewards obtained for Breakout, Qbert and SpaceInvaders using Dueling DQN with plasticity injection. The training loop was run for 10,000 episodes. A learning rate scheduler was used. The blue part is the reward value obtained from fixed training, while the yellow part is the reward value obtained from plastic training. During the plastic training, an $\epsilon$ value of 0 was used. The dashed lines are average rewards, which are computed for the fixed training rewards and the plastic training rewards separately.}
    \label{fig:ddqn_with_plasticity_0_eps_rewards}
\end{figure}

An important detail is that during the plastic training phase, a small amount of random actions were permitted ($10\%$). If the actions with the highest Q-values are only permitted, then only the strong connections get stronger over time, resulting in very bad overfitting. This is illustrated in figure \ref{fig:ddqn_with_plasticity_0_eps_rewards}, where the reward values get fixed at a certain value after some oscillations in the plastic training phase.

\begin{figure}[h]
    \centering
    \includegraphics[scale=0.6]{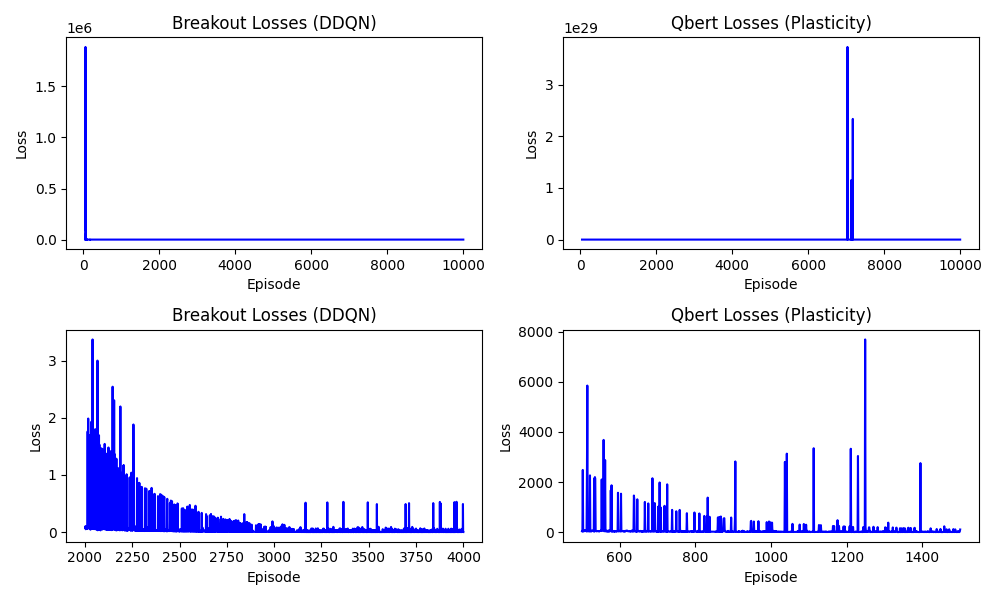}
    \caption{The loss values for Breakout and Qbert over the episodes. The loss values of Breakout was obtained from Dueling DQN training, while the loss values of Qbert was obtained from Dueling DQN training with plasticity. The below images are zoomed in segments of the top graphs.}
    \label{fig:losses}
\end{figure}

Figure \ref{fig:losses} illustrates the loss plots over the episodes. The loss curve for Breakout was taken from raw DQN training. In this plot, the loss value starts with a very high value and then seemingly becomes flat. But a zoomed-in version is shown in the bottom-left plot. It shows that the loss value maintains a decreasing trend with a few occasional overshoots. Qbert loss curve is taken while training with plasticity. It remains flat before the plastic training phase starts. At the start of the plasticity training, the loss value suddenly overshoots, oscillates for a while, and then again returns to the decreasing trend. The plot in the bottom right illustrates the decreasing trend by zooming into the graph.

\begin{figure}[h]
    \centering
    \includegraphics[scale=0.47]{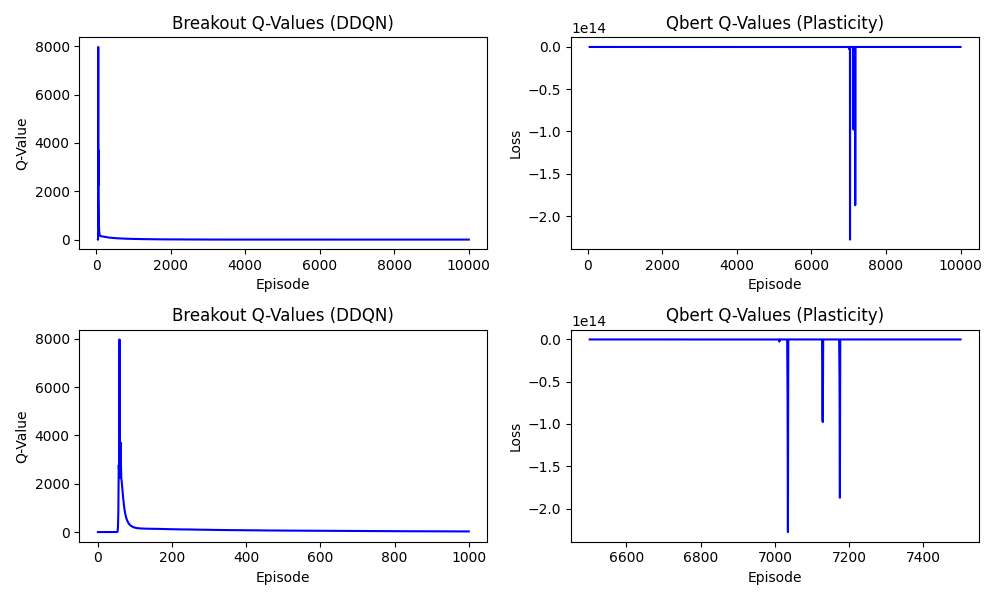}
    \caption{The q-values for Breakout and Qbert over the episodes. The q-values of Breakout were obtained from dueling DQN training, while the loss values of Qbert were obtained from Dueling DQN training with plasticity. The below images are zoomed in segments of the top graphs.}
    \label{fig:q_values}
\end{figure}

Figure \ref{fig:q_values} plots the maximum Q-values over the episodes. Breakout Q-values are taken from the DDQN training. It shows that the highest Q-value increases at the very beginning and then falls drastically over a few episodes and then remains flat. For Qbert which was taken from the plastic DQN training, at the beginning of the plastic phase, the max Q-value suddenly falls below zero, oscillates for a short time, and then comes back to a fixed level. This doesn't agree with the Q-value plots shown in Minh et al. [7], which shows that the max Q-values always keep an increasing trend This can be due to the fact that this work computes the max Q-values differently than the referenced paper.

\begin{figure}[H]
    \centering
    \includegraphics[scale=0.55]{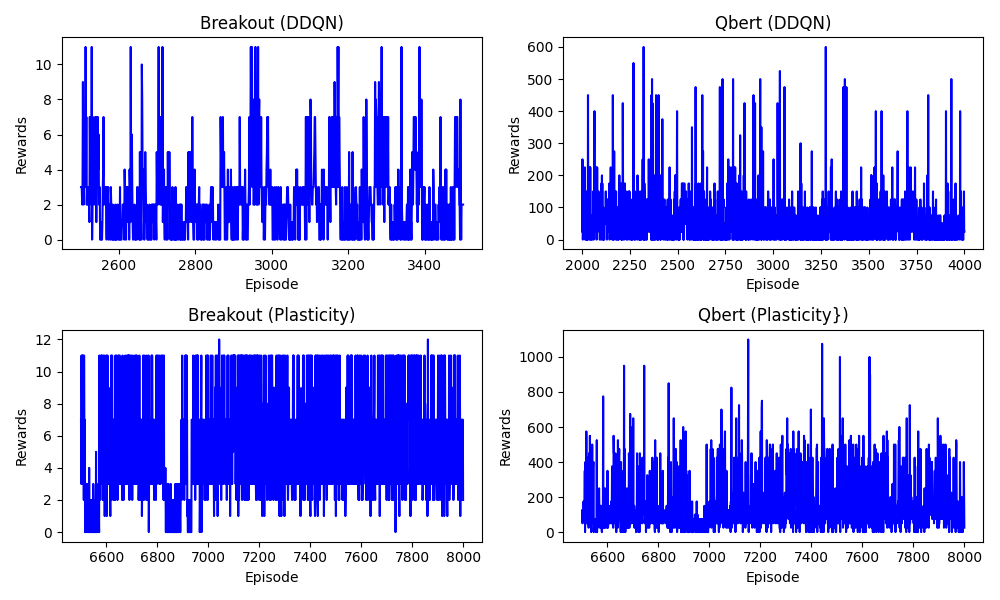}
    \caption{The zoomed-in reward values for Breakout and Qbert. The top segments are obtained by dueling DQN training, while the bottom two segments were obtained by dueling DQN training with plasticity injection.}
    \label{fig:rewards_zoomed_in}
\end{figure}

In all of the reward graphs shown, the reward values go up and down seemingly randomly. It seems like the system learns how to play the game well and then forgets after a while. This can be discussed more in the zoomed-in versions of the reward graphs in Figure \ref{fig:rewards_zoomed_in}. In all of these graphs, the reward values increase, stabilize for a while, and then suddenly fall. This phenomenon is called "catastrophic forgetting," and this is a common problem in reinforcement learning [16]. In the problem of this paper, the experiences are correlated and very similar. Training such systems with gradient descent is not ideal. Using a large replay buffer with priority sampling might reduce this issue. Interestingly, in the plasticity training graph snippets, the values are much more stable compared to the dueling DQN snippets. Hence, combined with other solutions, plasticity training might be able to stabilize reinforcement learning.

\begin{figure}[h]
    \centering
    \includegraphics[scale=0.9]{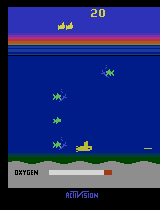}
    \caption{A game frame from Seaquest. The spaceship should turn left, but it's still firing, hinting at the model was overfit.}
    \label{fig:overfitted_game_frame}
\end{figure}

In multiple places above, it was mentioned that our model might be overfitted. Let's see this practically. In shooter games like SeaQuest, rewards are obtained when a bullet hits an enemy spaceship. So rewards are immediately awarded after appropriate shooting actions. Now the model might learn that shooting generates reward and triggers the shooting action more and more, ignoring the movement actions. In this way, the replay buffer will be filled with experiences containing the shoot action; hence, random sampling will mostly pick these actions, and the model will be repeatedly optimized with these strengthening the corresponding connections. With this problem, training for longer even makes the overfitting worse. By using a large replay buffer with priority sampling, the model will pick up the experiences with the highest loss values, and gradually it will evict the experiences with the lowest loss values; hence, actions other than shooting will be more likely to be picked up. Moreover, a better discounting strategy will help the model to understand the credits of subsequent good actions, not just the terminal ones. In Figure \ref{fig:overfitted_game_frame}, an enemy ship is just on the left of the agent's ship, so the agent should move left while it's just doing the overfitted action, shooting.

\subsection{Comparision}

\begin{center}
\begin{tabular}{|l|r|r|r|r|r|r|r|}
\hline
& Breakout & Q*bert & Seaquest & S. Invaders \\
\hline
Random [7] & 1.2 & 157 & 110 & 179\\
Sarsa [19] & 5.2 & 614 & 665 & 271\\
Contingency [20] & 6 & 960 & 723 & 268\\
DQN [7] & 168 & 1952 & 1705 & 581\\
Human [7] & 31 & 18900 & 28010 & 3690\\
HNeat [8] & 52 & 1325 & 800 & 1145\\
Our DQN & 13 & 1325 & 720 &1065\\
DQN with Plasticity & 15 & 1150 & 600 & 965\\
\hline
\end{tabular}
\label{table:comparision}
\end{center}

Table \ref{table:comparision} shows the comparison of our methods with the existing Atari learning methods. While our method performs better than the older methods [19, 20] and is almost similarly performant to HNet [8], it's much less performant than DeepMind's DQN method. This is expected due to the fact that in this work, priority sampling is not incorporated, which will be a game changer in this problem. Moreover, DeepMind used a sophisticated training infrastructure and trained for a very long time, whereas this work is trained on a base MacBook Pro. Finally, this is our first work in this area, and with a few iterations, we can make our method better over time.

\section{Limitation and Future Scope}

\subsection{Highlights}

The main highlights from the results are that after the initial training, the plastic weights can be trained lifelong, and this ongoing training helps improve performance, such as the accumulation of rewards, stability, etc. From the results, it is clear that during the plasticity training phases, the mean reward value is much higher and more stable than training using only DQN. The reason for this stability lies in the fact that the best weights, resulting in the highest scores, are restored and frozen while the plasticity contribution to the output is low. In this way, the system can preserve it’s best performance while remaining open to new experiences for life.\\

This work presents a new implementation platform to leverage reinforcement learning for solving complex problems with large state spaces. While such problems have been explored previously by large organizations like DeepMind, reimplementing their techniques and attempting new ideas alongside them (like the plasticity injection approach used here) can potentially lead to even better results. Irrespective of performance improvements, such works contribute to the advancement of reinforcement learning by introducing novel techniques for optimizing the algorithms.

\subsection{Future Scope}

\begin{itemize}

    \item The main limitation of this work is the diversity of the training samples. Currently, the system uses a FIFO queue of limited size, and the training batches are sampled randomly. If the neural network gets overfitted for a particular action, then the corresponding connections are strengthened over and over, regardless of how long the system is trained for. The biggest performance improvement can be achieved by using a priority replay buffer. A priority replay buffer intelligently samples the most important experiences from the replay buffer to train the neural network. This ensures that the network learns from a diverse set of experiences, reducing overfitting to a particular subset of the data. Additionally, a prioritized experience replay can lead to more efficient learning by increasing the probability of replaying important transitions that the network is currently struggling to learn. This will also improve the sampling efficiency of the network.\\

    \item Beside improving the sampling efficiency, the performance can possibly be improved by adding state and action transition effects to the Q-Learning formula. Currently, the system is fed with a sequence of states as history, but the formula only takes the last action into account. By feeding the system with the action sequence and discounting by a factor, the system can better learn the combined effect of a series of actions. To achieve this, the Q-learning formula might need to be adjusted accordingly. Furthermore, random sampling from the replay buffer can get experiences out of order, resulting in the current discounting mechanism having no effect. This aspect needs to be fixed in subsequent implementations.\\

    \item Instead of using hard prediction, where the model selects the action with the highest Q-value, an alternative approach is to use a soft prediction method. In this approach, the Q-values are converted into probabilities, and the next action is selected randomly based on these probabilities. This method encourages exploration by allowing the model to occasionally choose actions other than the one with the highest Q-value, while still prioritizing the actions with higher Q-values. By promoting exploration, the model may discover better strategies and avoid getting stuck in local optima.\\

    \item  The performance of the system can be improved by using an ensemble of multiple models. An ensemble method combines the predictions from multiple models, which can lead to better generalization and robustness compared to a single model. There are several ways to create an ensemble, such as training multiple models with different initializations, architectures, or hyperparameters, and then combining their predictions through techniques like averaging, voting, or stacking. Ensembles can help reduce overfitting, capture more complex patterns in the data, and improve overall accuracy and stability.\\

    \item On the plasticity aspect, further exploration can be done by optimizing the hyperparameter values. Different values of plasticity hyperparameters $\alpha$ and $\eta$ can be tried to find the best combination. Furthermore, other complex plasticity update rules can be tried, such as Oja’s rule [9]. A separate neural network can be used as the source of plastic weights to see how it performs instead of simple Hebbian update rules.\\

    \item Finally, the system needs to be trained for longer on powerful hardware. Currently, the training is done on a laptop with limited resources; hence, only a limited number of episodes could be run for each environment. These systems need to be trained for a very long time to get good results. Also, a large replay buffer should be used to get a diverse range of experiences. Running on a laptop prevents the use of a large replay buffer.

\end{itemize}

\section{Conclusion}

This project explored the implementation of various deep reinforcement learning techniques for learning complex tasks, such as playing games. As an extension, differentiable plasticity was incorporated to enable continued training of the neural networks after their initial training phase. The integration of plasticity demonstrated significant improvement in stability and performance compared to normal training over the same number of episodes. This improvement suggests that plasticity injection could be a promising direction for optimizing reinforcement learning algorithms. While the results indicate further scope for enhancements in the reinforcement learning aspects, the benefits of plasticity implementation are evident. Ultimately, several avenues for future improvements were identified, paving the way for continued advancements in this field.

\section*{Code Availability}

The code for the experiments are available in this git repository, https://github.com/ashfaq1701/dqn-atari.

\section*{References}

\begin{enumerate}
    \item R. Sutton and A. Barto, \emph{Reinforcement learning: An introduction}, MIT press, 1998.
    
    \item A. Salehin, \emph{Utilizing Discounted Policy Gradients to Learn How to Play Atari Games}, Adaptive Systems assignment, University of Sussex, 2024.
    
    \item R. Bellman, "A Markovian decision process," in \emph{Indiana Univ. Math. J.} 6, 1957.
    
    \item C. Watkins and P. Dayan, "Q-learning," in \emph{Machine learning}, 1992, 279-292.
    
    \item H. Hasselt et al., \emph{Deep Reinforcement Learning with Double Q-Learning}, in \emph{Proceedings of the 30th AAAI Conference on Artificial Intelligence}, 2015, 2094–2100.
    
    \item Z. Wang et al., \emph{Dueling Network Architectures for Deep Reinforcement Learning}, in arXiv preprint arXiv:1511.06581, 2015.
    
    \item V. Mnih et al., \emph{Playing Atari with Deep Reinforcement Learning}, in arXiv preprint arXiv:1312.5602, 2013.
    
    \item V. Mnih, K. Kavukcuoglu, D. Silver, et al., \emph{Human-level control through deep reinforcement learning}, in \emph{Nature} 518, 2015, 529–533.
    
    \item T. Miconi, J. Clune and K. Stanley, \emph{Differentiable plasticity: training plastic neural networks with backpropagation}, in \emph{Proceedings of Machine Learning Research}, 2018.
    
    \item R. Williams, \emph{Simple Statistical Gradient-Following Algorithms for Connectionist Reinforcement Leaning}, in \emph{Machine Learning} 8, 1992, 229–256.
    
    \item A. Samuel, \emph{Some studies in machine learning using the game of checkers}, in \emph{IBM Journal of Research and Development}, 1959.
    
    \item D. Silver et al., \emph{Mastering the game of Go without human knowledge}, in \emph{Nature}, 2017.
    
    \item Tom Schaul et al., \emph{Prioritized Experience Replay}, in arXiv preprint arXiv:1511.05952, 2015.
    
    \item Matteo Hessel et al., \emph{Rainbow: Combining Improvements in Deep Reinforcement Learning}, in arXiv preprint arXiv:1710.02298, 2017.
    
    \item D. Hebb, \emph{The organization of behavior: A neuropsychological theory}, Wiley, 1949.
    
    \item A. Geron, \emph{Hands on machine learning with Scikit-Learn and Tensorflow}, O’Reilly, 2022, 1044-1099.
    
    \item J. Mouret and P. Tonelli, \emph{Artificial Evolution of Plastic Neural Networks: a few Key Concepts}, in \emph{Studies in Computational Intelligence}, 2015.
    
    \item A. Joghataie and O. Torghabehi, \emph{Simulating Dynamic Plastic Continuous Neural Networks by Finite Elements}, in \emph{IEEE Transactions on Neural Networks and Learning Systems}, 2014.
    
    \item M. Bellemare, Y. Naddaf, J. Veness and M. Bowling, \emph{The Arcade Learning Environment: An Evaluation Platform for General Agents}, in \emph{Journal of Artificial Intelligence Research}, 2012.

    \item M. Bellemare, J. Veness and M. Bowling, \emph{Investigating contingency awareness using atari 2600 games}, in \emph{AAAI}, 2012.

    \item M. Hausknecht, R. Miikkulainen and P. Stone, \emph{A neuro-evolution approach to general atari game playing}, 2013.
\end{enumerate}

\end{document}